\documentclass[twocolumn,letterpaper]{ieeeconf}

\usepackage{amsmath}
\usepackage{amssymb}
\usepackage{graphicx}
\usepackage{subfigure}
\usepackage{verbatim}
\usepackage[thinlines,thiklines]{easybmat}
\usepackage{latexsym}
\usepackage{cite}
\usepackage{stmaryrd}
\usepackage{booktabs}
\usepackage{multirow}
\usepackage{multicol}
\usepackage{threeparttable}
\usepackage{textcomp}
\usepackage[ruled]{algorithm2e}
\usepackage{diagbox}
\usepackage{array}
\usepackage{color,soul}
\sethlcolor{yellow} 

\newcolumntype{M}[1]{>{\centering\arraybackslash}m{#1}}
\newcolumntype{N}{@{}m{0pt}@{}}

\usepackage[left=0.75in,top=0.75in,right=0.75in,bottom=0.75in]{geometry}
\newbox\tempbox

\def\sign{\mathop{\rm sign}\nolimits}

\begin{document}

\title{\bf \Large A Reduced-Order Resistive Force Model for Robotic Foot-Mud Interactions}

\author{Xunjie Chen, Jingang Yi, and Jerry Shan\thanks{X. Chen, J. Yi, and J. Shan are with the Department of Mechanical and Aerospace Engineering, Rutgers University, Piscataway, NJ 08854 USA (email: xc337@rutgers.edu, jgyi@rutgers.edu, jshan@soe.rutgers.edu).}}

\maketitle

\begin{abstract}
Legged robots are well-suited for broad exploration tasks in complex environments with yielding terrain. Understanding robotic foot-terrain interactions is critical for safe locomotion and walking efficiency for legged robots. This paper presents a reduced-order resistive-force model for robotic-foot/mud interactions. We focus on vertical robot locomotion on mud and propose a visco-elasto-plastic analog to model the foot/mud interaction forces. Dynamic behaviors such as mud visco-elasticity, withdrawing cohesive suction, and yielding are explicitly discussed with the proposed model. Besides comparing with dry/wet granular materials, mud intrusion experiments are conducted to validate the force model. The dependency of the model parameter on water content and foot velocity is also studied to reveal in-depth model properties under various conditions. The proposed force model potentially provides an enabling tool for legged robot locomotion and control on muddy terrain.
\end{abstract}

\section{Introduction}
\label{Sec:intro}

Although legged robots are attractive and well-suited for exploration tasks, yielding terrain such as granular or muddy ground brings a significant challenge for mission operations and field deployment. It is challenging to guarantee the rapid and safe locomotion of legged robots on deformable and multifaceted terrains~\cite{AguilarRPP2016,godon2023maneuvering}. Recent study of robot locomotion has mainly focused on granular terrain (e.g., sand) using the resistive force theory (RFT)~\cite{li2013terradynamics,zhang2014effectiveness,treers2021granular,huang2022dynamic}, digging/burrowing strategies~\cite{hosoi2015beneath,jung2011dynamics,huh2023walk} and walking locomotion~\cite{xiong2017stability,chen2023energy,chen2024ICRA}. However, legged locomotion on yielding terrain such as muddy ground is rarely  investigated. It is urgently needed to extend and build the force laws/models from dry granular media to wet and heterogenous media such as mud.

Although no force model was reported in~\cite{bagheri2017animal,bagheri2023mechanics}, wheel-legged robot locomotion experiments were conducted on wet granular materials to show how water saturation level affected robot motion. In~\cite{ma2022granular}, an extended RFT model was built on piecewise functions and used for saturated wet sand locomotion. Experiments were presented to validate the model. In~\cite{hossain2020drag}, an immersed pulling scenario through wet granular packing was considered. The Darcy-flow mechanism and a one-dimensional visco-elastic-plastic drag-force model were applied to interpret experimental observations and infer underlying physics.

Unlike granular materials, mud rheology is highly sensitive to temperature, clay type, and solid concentration level. It is challenging to predict the foot-mud interaction force from its ingredient components~\cite{coussot1994behavior}. Moreover, mud substrates interact with each other with the presence of water, producing attracting force and changing mud rheology (e.g., cohesion and suction) significantly. Rigorous constitutive models for mud rheology such as Herschel-Bulkley (HB) models~\cite{herschel1926measurements,bocquet2009kinetic} were used to describe mud flow curves, namely, the relationships among shear stress and shear rate. A new contribution of constitutive equation in\cite{caggioni2020variations} was proposed to combine “Bingham” model\cite{bingham1922fluidity} and the model in \cite{bocquet2009kinetic}. Another new viscoelastic thixotropic model was developed for a scalar form of mud rheology~\cite{ran2023understanding}. A parallel combination of a infinite shear viscosity damper plus a viscoelatic Maxwell model was used to explain mud rheology.

The above-mentioned fundamental constitutive models in~\cite{herschel1926measurements,bocquet2009kinetic,bingham1922fluidity,caggioni2020variations,ran2023understanding} cannot be directly applied to study robotic-foot/mud interactions and their corresponding mud rheological responses. Legged flipper robots were previously studied on muddy terrain~\cite{liang2012amphihex,ren2013experimental,liu2023adaptation}. In~\cite{liu2023adaptation}, two featured locomotion failure mechanisms were discussed and related to mud water content. The modeling approach used two simple thrust and drag force models in horizontal and vertical directions, respectively. Several helpful insights of stepping interaction on mud were presented for force hysteresis, suction force as well as energy consumption perspectives given differen mud conditions (e.g., water content)~\cite{godon2022insight}. However, the qualitative relationships between peak (suction) force and stepping locomotion velocity, water content, and foot shapes are still unclear. It is also challenging to estimate reaction forces by only sensing robot locomotion information (e.g., velocity and acceleration).

We present a resistive-force model for mud terrain interactions in robotic locomotion. We first conduct one-dimensional (1D) mud intrusion experiments to obtain mud-reaction-force characteristics such as relaxation time, cohesive suction force, and hysteresis. A reduced-order model is then proposed using a visco-elasto-plastic mechanism. The model considers both dynamic intrusion and withdraw (when suction happens) processes given a robot locomotion input. We carry out a series of experiments to estimate the model parameters and validate the model accuracy. The dependency of the model parameters on the locomotion velocity and mud properties (i.e., water content) is also discussed.

The main contribution of this work are twofold. First, the proposed resistive-force model for robotic-foot/mud interactions is new. The model directly predicts mud rheological response rather than through solving constitutive equations and therefore, enables potential use to develop a real-time force estimation and robot control with motion-sensing information. Second, this model presents a uniform, compact formulation for both the foot intrusion and withdrawn motions, describing the force-generation mechanism and including mud cohesion/suction with only a few parameters. These features are attractive for further robot dynamics integration and robot control.

The reminder of this paper is organized as follows. We introduce the experimental setup and basic findings on foot-mud interactions in Section~\ref{Sec:mudInteractions}. Details of the proposed resistive-force model are discussed in Section~\ref{Sec:RRFM}. Model validation experiments and discussion are presented in Section~\ref{Sec:results}. Finally, we summarize the conclusions and briefly discuss potential future research directions in Section~\ref{Sec:conclusions}.
\vspace{-1mm}
\section{One-Dimensional Foot-Mud Interactions}
\label{Sec:mudInteractions}

\subsection{Experimental Setup}
\label{subSec:ExpSetup}

We focus on one-dimensional mud-resistive-force modeling, and therefore, a series of 1D mud-penetration experiments were specifically performed with a prescribed mud formulation and experimental protocols.

\subsubsection{Materials}

\begin{figure}[h!]
  \centering
  \includegraphics[width=3.0in]{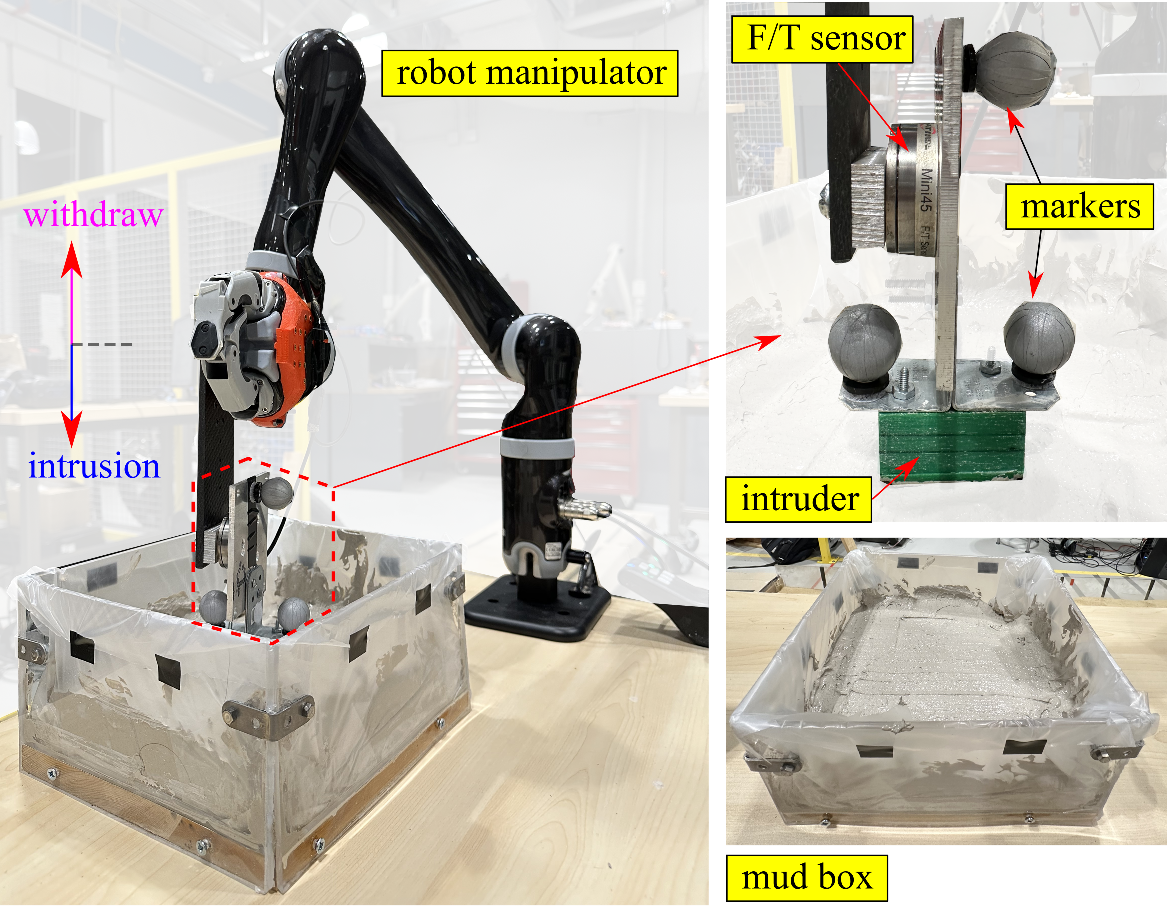}
  \caption{Mud-intrusion experimental setup with a cuboid intruder.}
  \label{fig:expSetup}
\end{figure}

Fig.~\ref{fig:expSetup} shows the experimental setup. All mud intrusion experiments were conducted within a container with a size of $28\times23\times 15$~cm. We filled the container with customized mud mixtures. The mixed mud materials included controlled volume of sand ($150$-$600$~$\mu$m graded standard silica sand from Gilson Company Inc.), clay (SP648 Sea Mix 6 from Seattle Pottery Supply), and water in specified proportion. For all reported data, we first set the clay-to-sand ratio as $3$:$1$ in all cases. We used volume ratio to define the water content of muddy mixtures, that is, $W=V_w/\left(V_c+V_s+V_w\right)$, where $V_c$, $V_s$, and $V_w$ present the occupied volumes of clay, sand, and water, respectively. The synthetic-mud  properties were achieved by fully mixing ingredients to make the mud dynamicallt similar to natural mud~\cite{kostynick2022rheology}. By changing the water content $W$, both mechanical (related to yield stress) and rheological (related to deformation) properties of mud can be controlled. We used a 3D-printed cuboid (with size of $51\times 38 \times 25$~mm) as the intruder for all intrusion tests.

\subsubsection{Experimental protocol}

To implement 1D motion along the vertical direction, the intruder was mounted at the end-effector of a robotic manipulator (Jaco from Kinova Inc.) to move downward and upward. Optical markers and a motion-capture system (10 Bonita cameras from Vicon Ltd.) were used to obtain the real motion of the intruder. A 3-axis force/torque (F/T) sensor (model mini45 from ATI Inc.) was mounted between the intruder and the manipulator's end-effector to measure the resistive forces during the penetration motion.

For a single intrusion test under a certain water content, the intruder first moved downward and vertically with a prescribed constant velocity until reaching at a certain designed intrusion depth. We then maintained the intruder position temporarily to let the reaction force become stable within a time interval (6s). Finally, we withdrew the intruder upward with the same velocity as the intrusion process until it totally separated away from clinging mud. Both force and motion data were synchronized and recorded with the $100$~Hz sampling frequency during the three stages of movement described above. For each water-content level and intrusion/withdrawing velocity condition, three trials were repeated, with the mud surface flattened before each trial.

\begin{figure*}[t!]
	\subfigure[]{
\hspace{-2mm}
		\label{fig:forceProfile_T}
\includegraphics[width=2.34in]{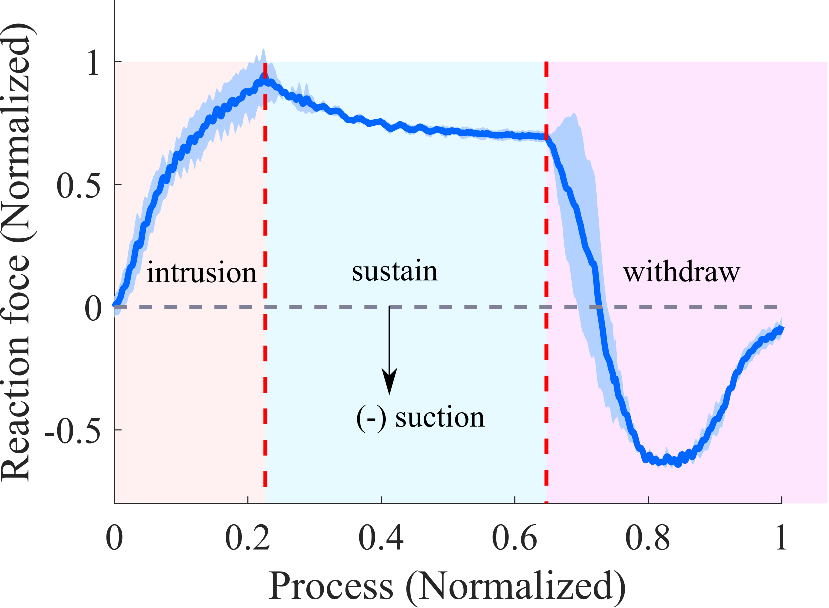}}
	\subfigure[]{
\hspace{-0mm}
		\label{fig:forceProfile_Z}
\includegraphics[width=2.33in]{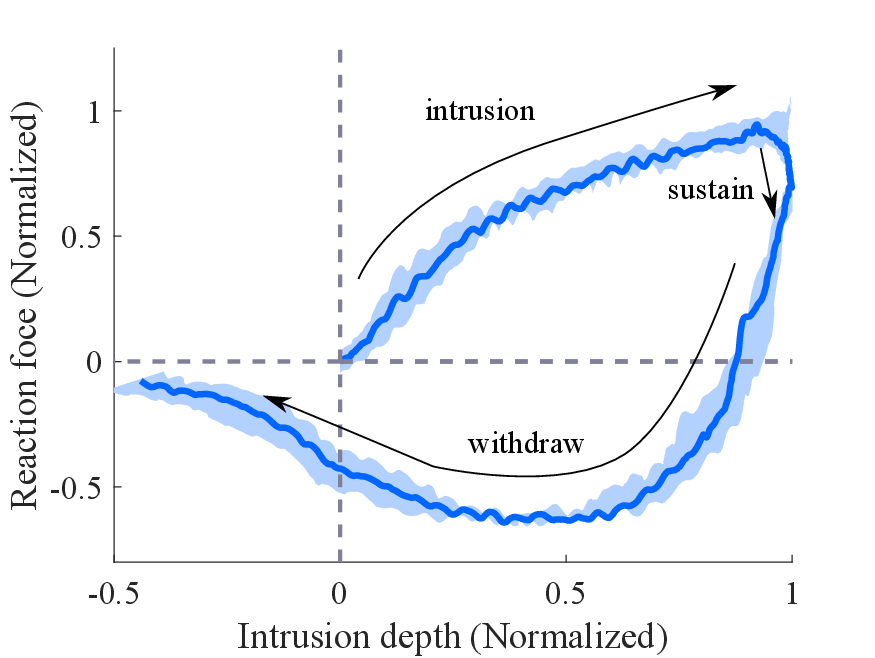}}
\hspace{-3mm}
	\subfigure[]{
		\label{fig:suctionView}
\includegraphics[width=2.1in, height=1.6in]{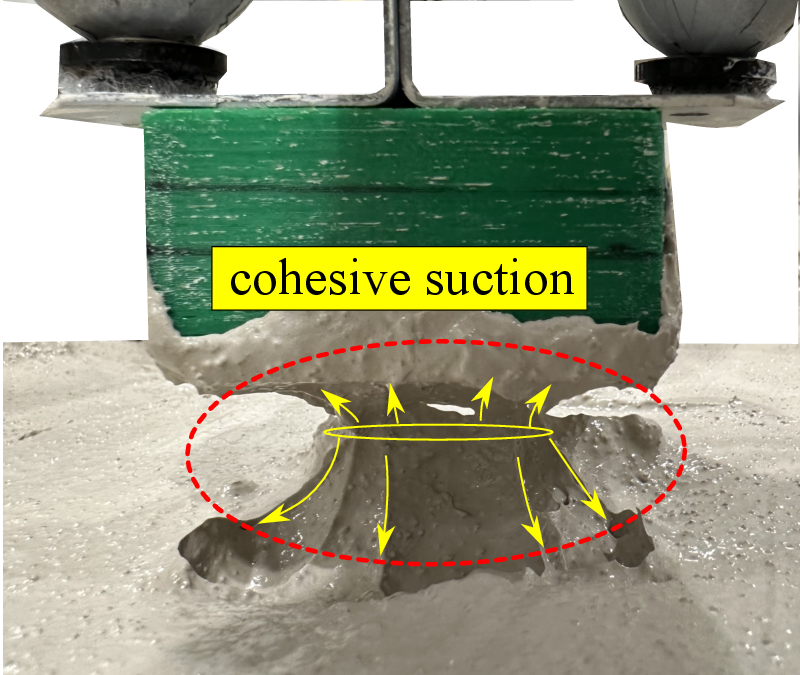}}	
\caption{A typical resistive-force profile (normalized). (a) In normalized process (time) domain. (b) In normalized intrusion-depth domain. Shaded regions demonstrate one-standard deviation from 3 trials. (c) The necking of mud that leads to a cohesive suction force during the withdraw process.}
	\vspace{-2mm}
\end{figure*}

It is noted that we intentionally did not immerse the intruder completely into the substrate because we tried to exclude any additional force due to any mud on the top. Through this motion, we captured the natural relaxation characteristics at a sustained intrusion depth, as well as emphasized the cohesive suction force between the mud and the bottom contact surface of the intruder upon withdrawal, rather than the resistive weight induced by materials on top. This is different from the experimental setup and configuration in~\cite{hossain2020drag,godon2022insight}.

\subsection{Force Profile}
\subsubsection{Nonlinear intrusion force}
Fig.~\ref{fig:forceProfile_T} shows one selected trial result of the reaction force in experiments. The reaction force magnitude was normalized by the maximum intrusion value while the process was normalized by total time duration. The reaction force experienced three stages: intrusion increasing, sustain, and withdrawing suction regimes. Fig.~\ref{fig:forceProfile_Z} shows the reaction force versus the intrusion displacement, which was normalized by the final depth. A nonlinear hysteresis was observed during entire intrusion and withdraw processes. Unlike linear relationship reported for granular materials in~\cite{qian2015principles,xiong2017stability}, intrusion resistive force in mud media increases nonlinearly with intrusion depth. Therefore, a simple linear depth-dependent force model as the RFT-based models in~\cite{li2013terradynamics,treers2021granular} cannot accurately describe the hysteresis phenomenon, and a nonlinear model is needed. It is also noted that the resistive force during the sustain part of the process decayed gradually to a stable value as shown in Fig.~\ref{fig:forceProfile_T} due to the natural visco-elasticity of mud under a given applied force (stress). A clear force (stress) relaxation was observed and this relaxation time is assumed to be only related to material characteristics such as clay-to-sand ratio and water content.

\subsubsection{Cohesive suction force and necking}

As the intruder withdrew, the reaction force dropped rapidly and crossed zero before becoming negative and creating a suction under the intruder. Fig.~\ref{fig:suctionView} illustrates the cohesive property and plastic deformation (necking) of mud. The necking effect provides a suction force pulling the intruder downwards, which is a particular feature for foot-mud interactions. As the intruder is pulled up further, the neck narrows and eventualy breaks and and the suction is lost, leading to zero resistive force. Generally, the suction drops if the internal stress goes beyond the fracture limit of the mud. Nevertheless, the pulling force (magnitude) decays slowly, which is attributed to a similar visco-elasticity property as mentioned in the sustain regime.

Based on aforementioned insights on mud-intrusion force profile, we conclude that nonlinear modeling should be considered for both intrusion and withdrawal processes. Moreover, the goal of this study is to develop a unified force model to cover all the three intrusion regimes. A physical model-based method is preferable here for its interpretability.

\section{Reduced-Order Mud Resistive Force Model}
\label{Sec:RRFM}

\subsection{Visco-elasto-plastic Modeling}

\begin{figure}[h!]
	\centering
	\subfigure[]{
		\label{fig:mudSchematics_a}
		\includegraphics[width=1.5 in]{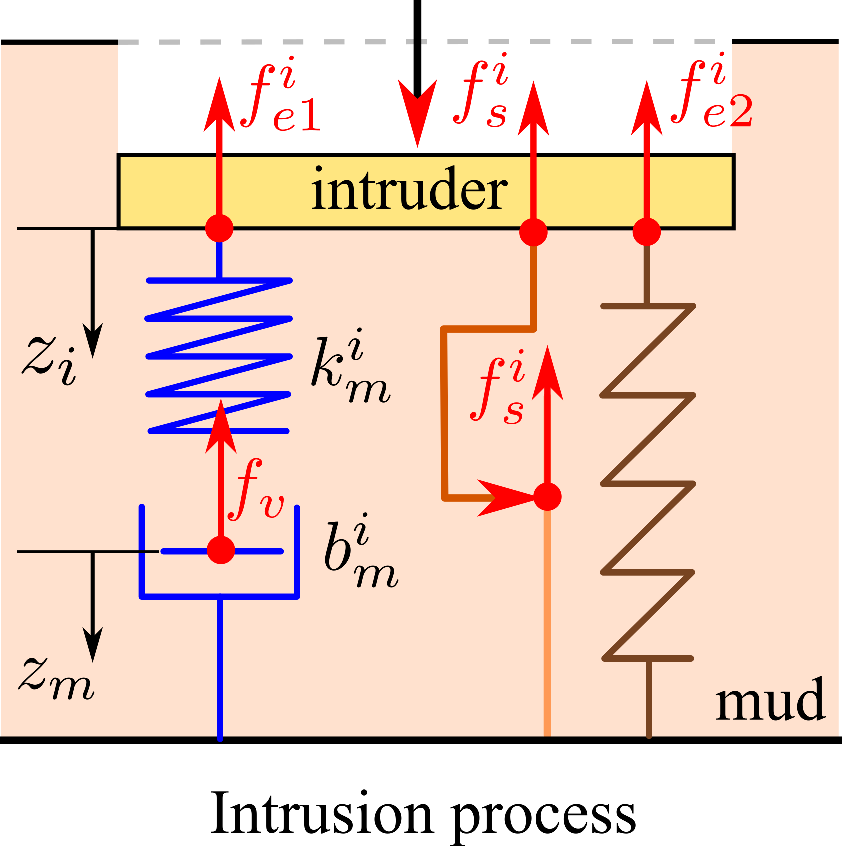}}
\hspace{2mm}
	\subfigure[]{
		\label{fig:mudSchematics_b}
		\includegraphics[width=1.5 in]{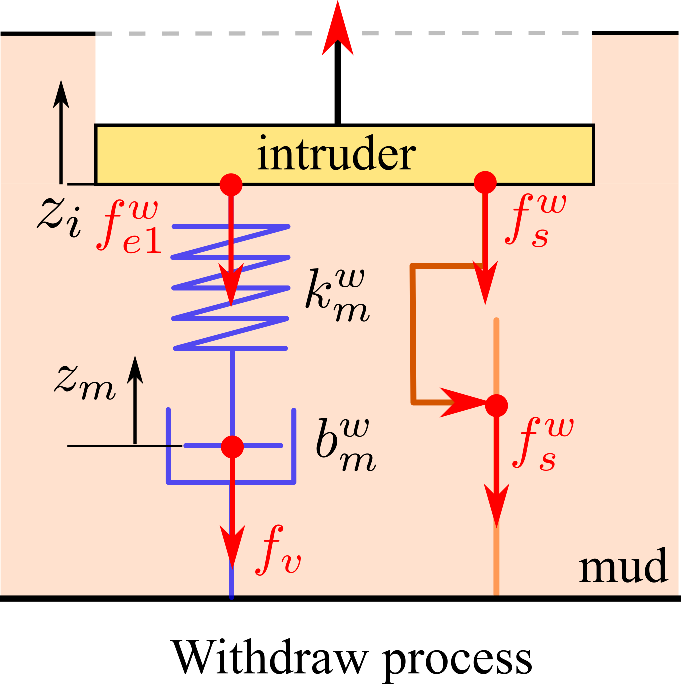}}
	\caption{Schematics of 1D foot/mud interaction models using combinations of visco-elasto-plastic elements for the (a) Intrusion and (b) Withdrawal process.}
\label{fig:mudSchematics}
	\vspace{-3mm}
\end{figure}

Instead of modeling mud complex physical behavior during the interaction, we present a reduced-order model based on a combination of elementary mechanism. A visco-elasto-plastic element of muddy material is considered here. Fig.~\ref{fig:mudSchematics} illustrates the schematics of the intruder-mud interaction modeling by a visco-elasto-plastic mechanism for both the penetration and withdrawal processes. We consider that the intruder moves only along the vertical direction, namely, penetrating (downward) and pulling (upward). Due to limited intrusion depth and mud compressibility and viscosity given a low water content, there is no mud  on the upper surface of the intruder. That means the resistive force only comes from the bottom interaction. The reaction force is considered uniform across the surface in contact with mud and then the total reaction force is calculated as $F=f_m S$ where $f_m$ is the resultant mud resistive force per area (i.e., stress) and $S$ is the total contact area.


\subsubsection{Intrusion modeling}

Fig.~\ref{fig:mudSchematics_a} illustrates the intrusion process when the intruder moves downward in mud. The model comprises a Maxwell visco-elastic element (i.e., a spring-damper system), a plastic slider, and another spring element that are connected in parallel. We infer the micro-elastic deformation of substrate as a visco-elastic element for the internal action by mud itself and its applied resistive stress to the intruder is denoted by $f_{e1}^i$. The single spring element denotes a resistive stress, denoted by $f_{e2}^i$, due to the macroscopic (bulk) deformation of the mud material enforced by the motion of the intruder. Finally, the additional slider contributes a viscous frictional stress, denoted by $f_s^i$.

By using the visco-elasto-plastic model, the resultant resistive stress applied to the intruder is calculated as
\begin{equation}
\label{eqn:RF_intrusion}
  f^i_m= f_{e1}^i + f_{e2}^i + f_s^i.
\end{equation}
Denoting the penetration displacement of the intruder originating from the mud surface as $z_i$ and is the internal displacement of mud substrates as $z_m$, we obtain the elastic stress $f_{e1}^i$ and viscous damping stress $f_v^i$ as
\begin{equation}
\label{eqn:f_e1}
  f^i_{e1} = k_m^i\left(z_i-z_m\right),\; f^i_v = b_m^i \dot{z}_m,
\end{equation}
where $k_m^i$ and $b_m^i$ represent the stiffness and damping coefficients for the mud, respectively. For the internal force balance of the visco-elastic element, we have $f^i_{e1}=f^i_v$ and therefore, from~\eqref{eqn:f_e1} we obtain
\begin{equation}
\label{eqn:spring_damper}
   b_m^i \dot{z}_m  + k_m^i z_m = k_m^i z_i.
\end{equation}
Given the motion input of the intruder $z_i$, from~\eqref{eqn:spring_damper} we update mud displacement $z_m$ to obtain the resultant visco-elastic force $f^i_{e1}$.

Considering the resistance of instantaneous volume change (decrease), the force $f^i_{e2}$ inferred by the single spring is obtained as
\begin{equation}
\label{eqn:f_e2}
 f^i_{e2} = \alpha \left(\frac{z_i}{H}\right)^{\beta},
\end{equation}
where $\alpha$ is defined as the stiffness related to direct volume change of mud, $\beta \in (0,1]$ is a constant, and $H$ is the intruder width. In general, we consider a nonlinear spring model for $f^i_{e2}$ given in~\eqref{eqn:f_e2}. Viscous friction force $f^i_s$ is considered as inertial drag at a regime where force magnitude increases quadratically with motion velocity~\cite{potiguar2013lift}. Therefore, we calculate the viscous friction force as
\begin{equation}
\label{eqn:f_s}
f^i_s = \sign(\dot{z}_i)\lambda \rho_m \dot{z}_i^2,
\end{equation}
where $\rho_m$ is the mass density of mud, $\lambda$ is scaling factor determined by calibration and function $\sign(x)=1$ for $x \geq 0$ and $-1$ otherwise.

\subsubsection{Withdrawing and beyond-necking modeling}

Suction force was observed during the withdrawal, as shown in the negative force in Fig.~\ref{fig:forceProfile_T}. We attribute this suction force to the cohesiveness of the mud as well as its viscosity due to the internal presence of water. However, as the displacement of mud increases, internal stress also increases until it reaches the yield stress, after which the neck fractures and the reaction force (magnitude) reduces to zero.

Fig.~\ref{fig:mudSchematics_b} shows the schematic of the withdrawing force model. We do not consider the resistance due to significant volume deformation, i.e., no nonlinear spring element as there was for the intrusion process. Therefore, the total reaction stress is computed as
\begin{displaymath}
f_m^w = f^w_{e1} + f^w_s,
\end{displaymath}
where we use superscript ``w'' to represent the force during the withdrawal process. The visco-elastic force $f^w_{e1}=k_m^w\left(z_i-z_m\right)$ and viscous friction force $f^w_s$ are calculated similar to~\eqref{eqn:spring_damper} and~\eqref{eqn:f_s}, respectively. Similar to~\eqref{eqn:spring_damper}, the stiffness and damping coefficients, denoted respectively by $k_m^w$ and $b_m^w$, are used such that $b_m^w \dot{z}_m  + k_m^w z_m = k_m^w z_i$ for the withdrawing process.

We denote the mud yield stress by $\sigma_y$. When the stress goes beyond $\sigma_y$, the mud neck break and the intruder separates from the bottom mud surface. This results in the drop of the suction-force magnitude shown in the later phase in Fig.~\ref{fig:forceProfile_T}. We consider that the displacement velocity decays to zero gradually instead of suddenly stopping deforming. To model this viscous behavior, a second-order filter is used to model the mud velocity after necking. The mud velocity follows that $V_m(s) = G_m(s) v_{m0}$, where $V_m(s)=\mathcal{L}(v_m(t))$ is the Laplace transformations of mud velocity $v_m(t)=\dot{z}_m(t)$ after necking and $v_{m0}$ is the mud velocity before necking. \textcolor{black}{Based on the physical mechanism analogue of the spring-damper system as shown in Fig.~\ref{fig:mudSchematics_b}, the lumped mud motion without considering intruder motion is $m \ddot{z}_m + b_m \dot{z}_m + k_m z_m = 0$. Considering initial conditions that $z_{m0}=0$ and $\dot{z}_{m0}=v_{m0}$ at necking and taking Laplace transformation, we obtain $Z_m(s) = \frac{mv_{m0}}{ms^2 + b_m s +k_m}$ and $V_m(s) = s Z_m(s) = \frac{ms}{ms^2 + b_m s +k_m}v_{m0}$.} Therefore, the filter takes the form of
\begin{equation}
\label{eqn:secondOrderFilter}
  G_m(s) = \frac{s}{s^2 + 2\zeta \omega_0 s + \omega_0^2},
\end{equation}
where both $\omega_0$ and $\zeta$ are constant parameters related to mud materials. Here $\omega_0 \propto \sqrt{k_m}$ and $\zeta \propto \frac{b_m}{\sqrt{k_m}}$. We set $v_{m0}(t)$ before necking as the command velocity input to $G_m(s)$ and step function that drops to zero is used for $v_{m0}(t)$ so that the mud velocity would settle down to zero with the initial velocity right before necking.

\subsection{Mud Resistive Force Model}

\begin{figure}[h!]
	\centering
	\includegraphics[width=3.1 in]{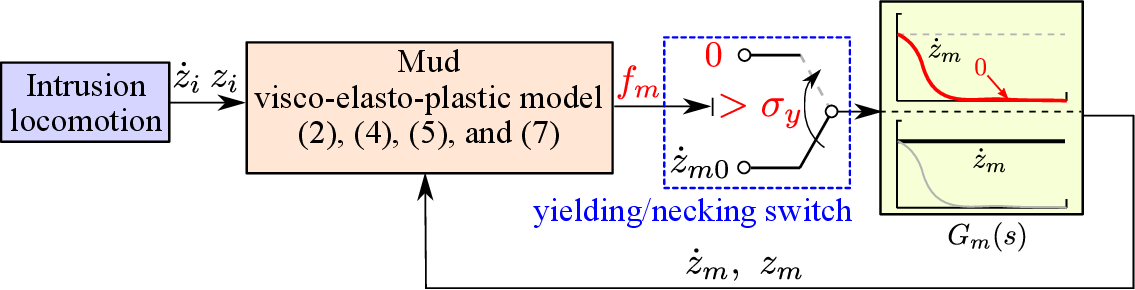}
	\caption{The mud resistive-force model diagram. A switch is used for regulating the mud velocity $\dot{z}_m$ to interpret necking after significant yielding.}
	\label{fig:modelPipeLine}
\vspace{-2mm}
\end{figure}

Fig.~\ref{fig:modelPipeLine} shows the flow chart of the proposed mud resistive force model. Based on the previous discussion of the visco-elasto-plastic model, we summarize the reduced-order mud resistive force that are given by the summation of the elastics and viscous forces in~\eqref{eqn:f_e1},~\eqref{eqn:f_s}, and~\eqref{eqn:f_e2}. The motion information of the intruder (i.e., $\dot{z}_i$ from motion capture system) is used as an input to model~\eqref{eqn:spring_damper} to estimate the mud displacement $z_m$ for force calculation. The unified resistive force is obtained as
\begin{equation}
\label{eqn:MudForceModel}
f_m^j = f^j_{e1} + f^j_s + \left(1-w(\dot{z}_i)\right)f^i_{e2}, \; j=i,w,
\end{equation}
where index $w(\dot{z}_i)=\frac{1}{2}\left[1-\sign(\dot{z}_i)\right]$ indicates whether the intruder moves downwards or upwards.

Under the withdrawing condition, we design a yielding/necking switch to regulate the mud displacement velocity by using the velocity filter as
\begin{equation}
\label{eqn:ifYielding}
  G_m(s) =
  \begin{cases}
    1, &|f^w_m| \leq \sigma_y \\
    \frac{s}{s^2 + 2\zeta \omega_0 s + \omega_0^2}, &|f^w_m| > \sigma_y.
  \end{cases}
\end{equation}
Without yielding/necking, the mud velocity after the filter holds the value before necking, that is, $\dot{z}_m = \dot{z}_{m0}$, where $\dot{z}_{m0}$ ($\dot{z}_{m}$) presents the mud velocity before (after) the filter. After yielding condition, the mud velocity is enforced to drop to zero, that is, $\dot{z}_m \to 0$ when $|f^w_m|>\sigma_y$. The values of the yielding strength $\sigma_y$, the filter parameters $\omega_0$ and $\zeta$, which depend on water content and mud compisition, are obtained by experimental identification.
\section{Model Validation and Results}
\label{Sec:results}

In this section, we first present the model calibration and then the validation results and discussion.

\renewcommand{\arraystretch}{1.2}
\setlength{\tabcolsep}{0.07in}
\begin{table*}[t!]
\centering
\caption{Parameters for mud with different water content and model estimation RMSE.}
\vspace{-1mm}
\label{tab:modelParas}
\begin{tabular}{ccccccccccc}
\toprule[1.2pt]
$W$ & $k_m^i$ [MPa/m] & $b_m^i$ [MPa/(m$\cdot s^{-1}$)] & $k_m^w$ [MPa/m] & $b_m^w$ [MPa/(m$\cdot s^{-1}$)] & $\alpha$ [MPa] & $\beta$ & $\sigma_y$ [KPa] & $\zeta$ & $\omega_0$ & RMSE [N]\\ \midrule[1.1pt]
$15\%$ & $1.21$ & $0.24$ & $1.48$ & $1.35$ & $0.17$ & $0.56$ & $17$ & $0.47$ & $4.09$ & $2.57$\\
$20\%$ & $0.70$ & $0.16$ & $1.71$ & $1.56$ & $0.12$ & $0.49$ & $14$ & $0.31$ & $3.45$ & $1.52$\\
$25\%$ & $0.26$ & $0.29$ & $1.21$ & $1.35$ & $0.04$ & $0.54$ & $6$ & $0.49$ & $2.23$ & $0.84$\\
$30\%$ & $0.11$ & $0.06$ & $1.27$ & $1.40$ & $0.01$ & $0.46$ & $2$ & $0.36$ & $1.44$ & $0.36$\\
$35\%$ & $0.28$ & $0.07$ & $1.16$ & $1.36$ & $0.01$ & $0.38$ & $2$ & $0.81$ & $2.21$ & $0.32$\\
\toprule[1.2pt]
\end{tabular}
\end{table*}

\setcounter{figure}{5}
\begin{figure*}[t!]
	\centering
	\subfigure[]{
		\label{fig:Result_force}
		\includegraphics[width=2.3 in]{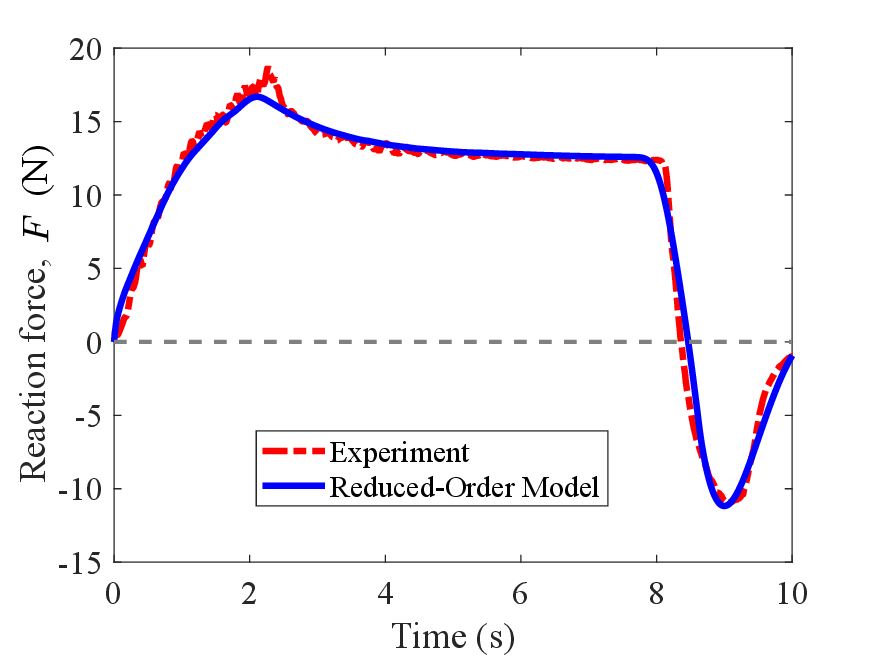}}
	\subfigure[]{
		\label{fig:Result_velocity}
		\hspace{-6mm}
		\includegraphics[width=2.1 in]{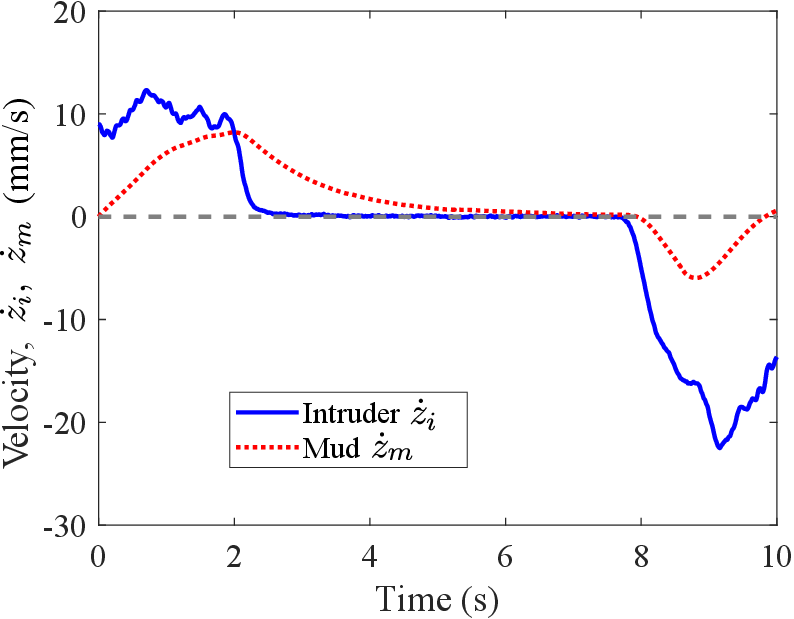}}
	\subfigure[]{
		\label{fig:Result_displacement}
		\includegraphics[width=2.1 in]{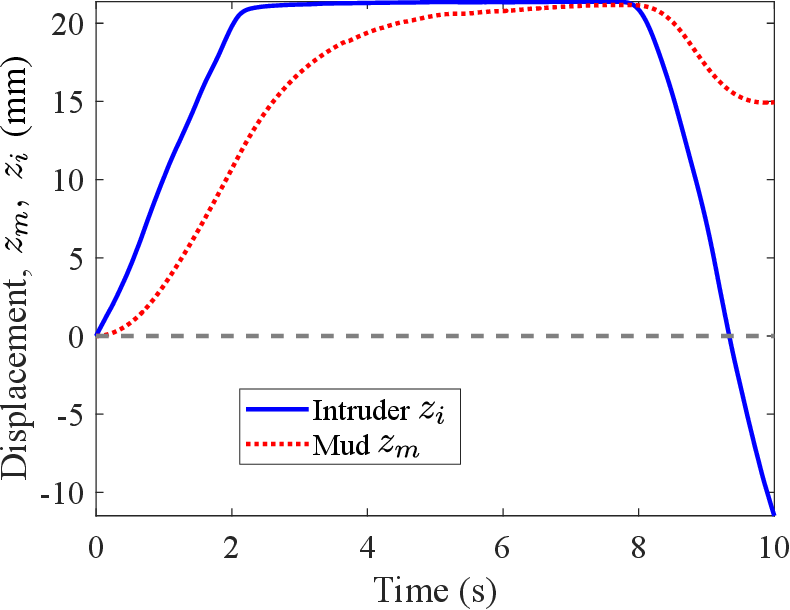}}
	\caption{Mud intrusion experiment and model validation results for the mud with water content $W=25\%$. (a) Reaction force experiments and comparison with model prediction. (b) Recorded intruder and mud velocity $\dot{z}_i$ and $\dot{z}_m$. (c) Intrusion displacements $z_i$ and $z_m$ for the intruder and the mud.}
\end{figure*}

\subsection{1D Intrusion and Withdrawing}

\subsubsection{Calibrations}

We first calibrated and estimated the mud parameters such as mud density through experiments. The mass density of the synthetic mud is $\rho_m = 1.84 \times 10^3$~kg/m$^3$ for mud water content $W=25\%$. It is found that as $W$ varied from $15\%$ to $35\%$ ($V_w$ was from 235~cm$^3$ to 600~cm$^3$), the mud density did not change much since the weight of water contributed not significantly to total mud materials. Therefore, we used the mud density of medium water content $W=25\%$ as a representation. Fig.~\ref{fig:lambda_Setup} illustrates the setup of the surface sliding experiments to obtain the scaling factor $\lambda$ in the viscous friction force~\eqref{eqn:f_s}. The sliding experiments were conducted by making the intruder contact the mud surface with a small depth and moving it along one direction. Three trials were repeated for each velocity condition. Fig.~\ref{fig:lambda_cali} shows the corresponding calibration results, which estimated the scaling constant to be $\lambda=0.013$.

\setcounter{figure}{4}
\begin{figure}[h!]
	\centering
	\subfigure[]{
		\label{fig:lambda_Setup}
		\includegraphics[width=2.2 in]{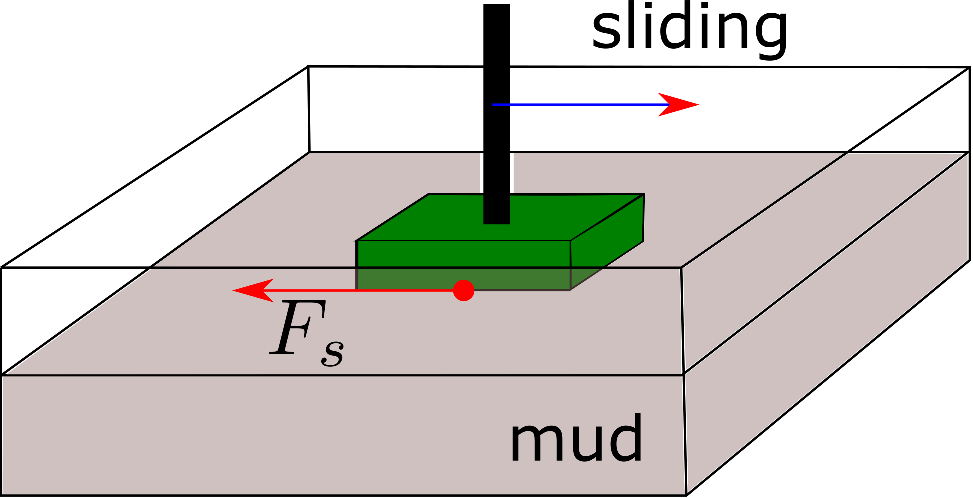}}
	\subfigure[]{
		\label{fig:lambda_cali}
		\includegraphics[width=2.9 in]{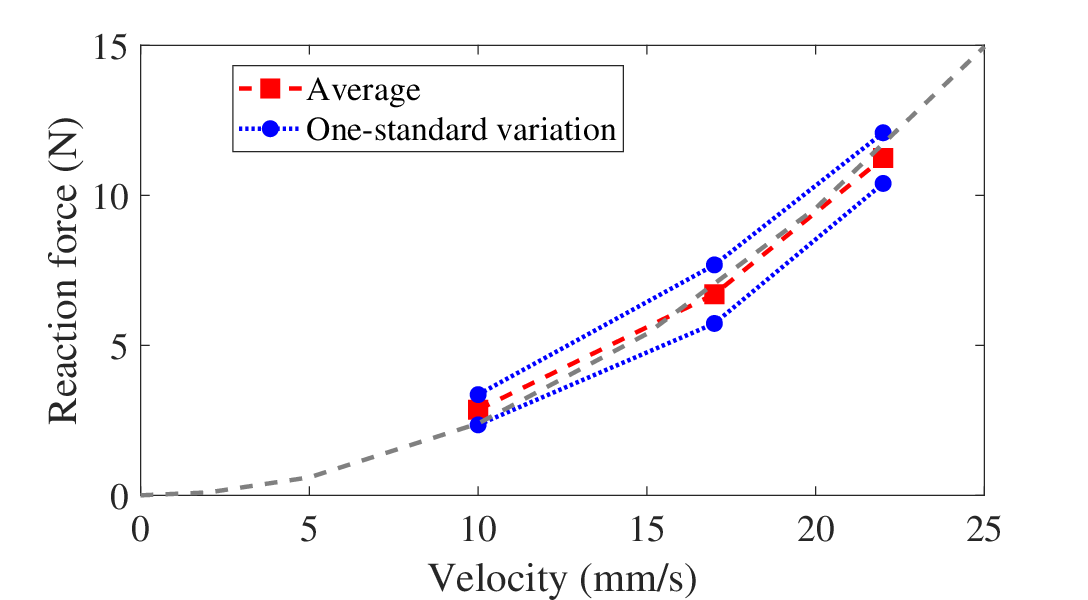}}
	\caption{Calibration of the scaling factor $\lambda$. (a) Sliding experiment setup schematics. (b) Calibration results.}
	\vspace{-2mm}
\end{figure}

In terms of the other model parameters, we conducted series of intrusion experiments under different water contents and moving velocities. Calibration was performed by formulating an optimization problem to minimize the difference between the model prediction and the experimental results. Table~\ref{tab:modelParas} lists the values of the key model parameters (i.e., $k_m^i$, $b_m^i$, $k_m^w$, $b_m^w$, $\alpha$, $\beta$, $\sigma_y$, $\zeta$, and $\omega_0$) that were obtained and estimated from the calibration process.

\setcounter{figure}{8}
\begin{figure*}[t!]
	\centering
	\subfigure[]{
		\label{fig:alphabeta_watercontent_a}
		\includegraphics[width=2.2 in]{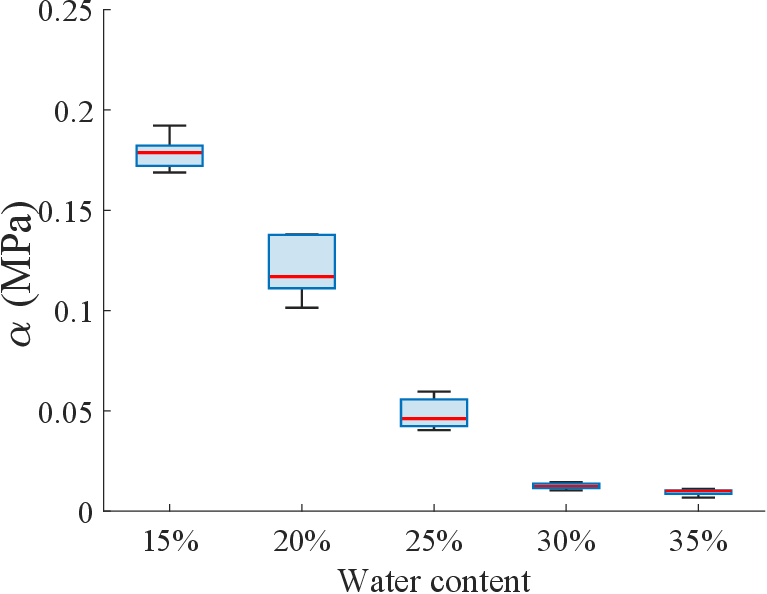}}
	\subfigure[]{
		\label{fig:alphabeta_watercontent_b}
		\includegraphics[width=2.15 in]{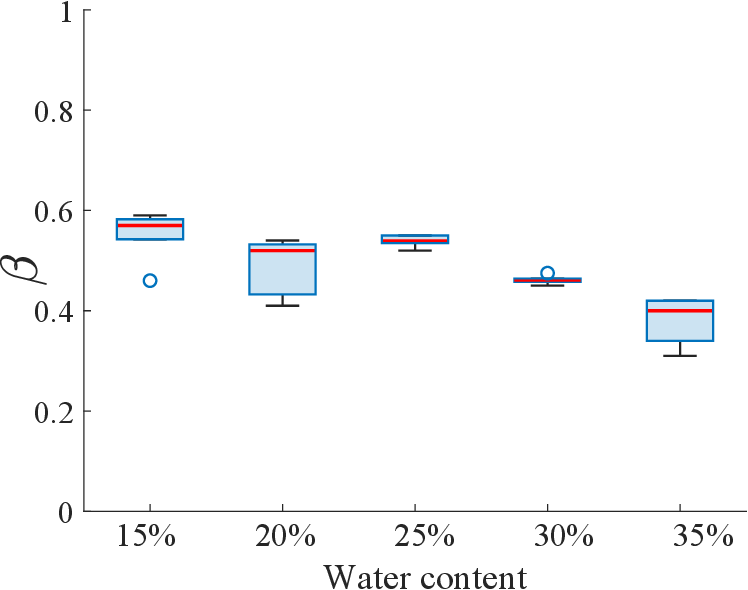}}
	\subfigure[]{
		\label{fig:alphabeta_watercontent_c}
		\includegraphics[width=2.25 in]{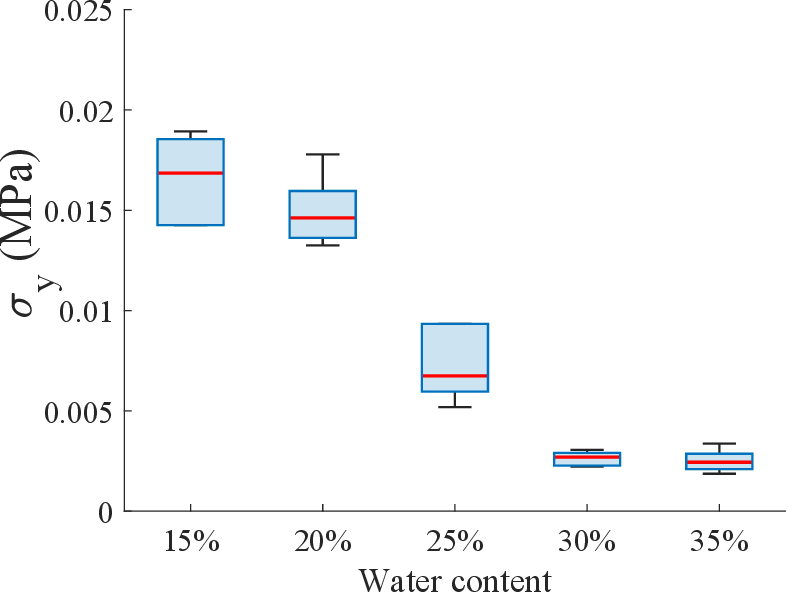}}
	\caption{The estimated parameter values for mude of various water content. (a) Stiffness parameter $\alpha$. (b) Stiffness parameter $\beta$. (c) Mud yielding threshold $\sigma_y$.}
\label{fig:alphabeta_watercontent}
\end{figure*}

\subsubsection{Validation results}

\setcounter{figure}{6}
\begin{figure}[!h]
  \centering
  \includegraphics[width=3.2in]{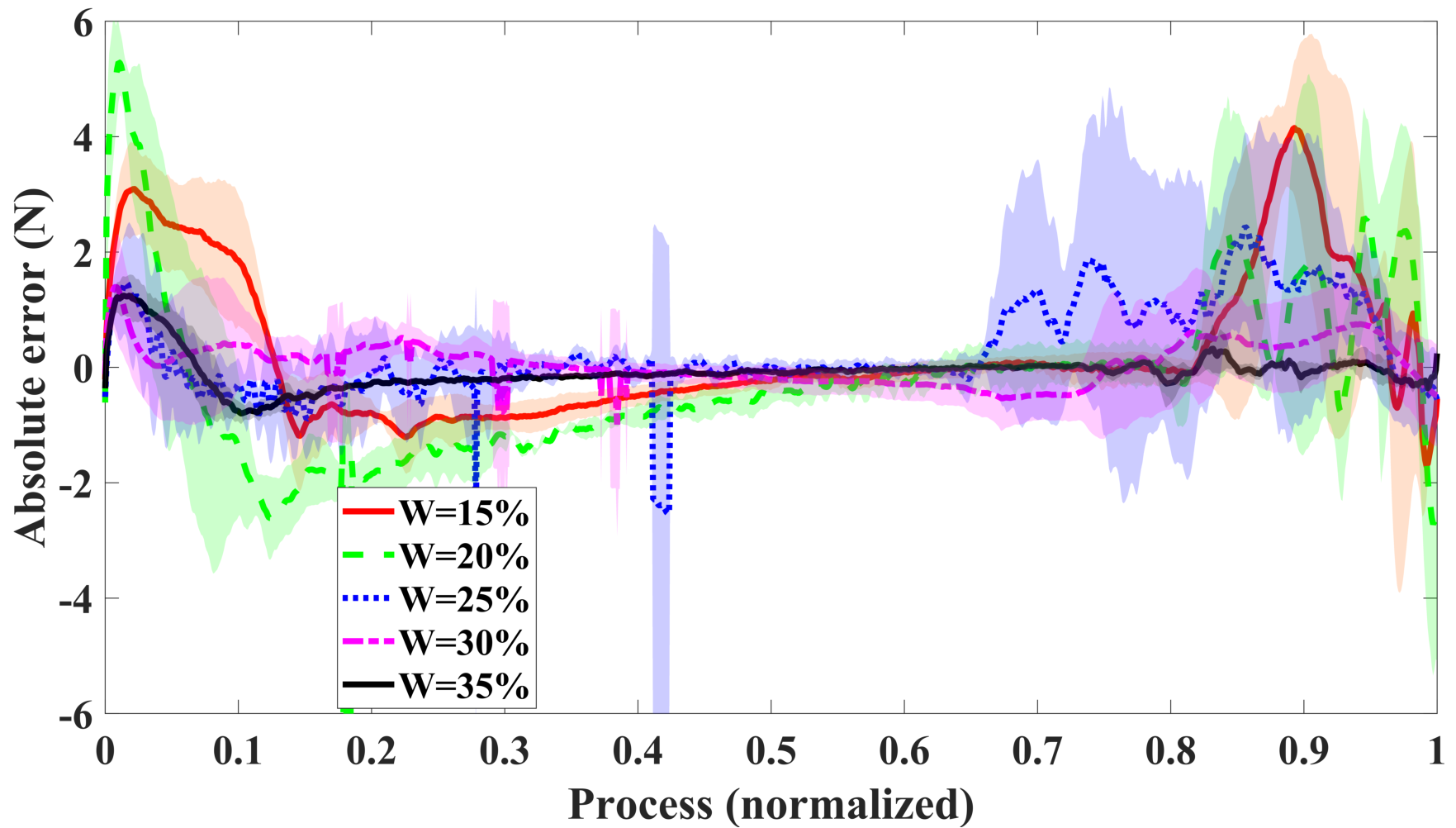}
  \caption{The estimation error profiles of the mud resistive force model under various water-content conditions.}
  \label{fig:Error}
\end{figure}

\begin{figure}[!h]
  \centering
  \hspace{-6mm}
  \includegraphics[width=3.1in]{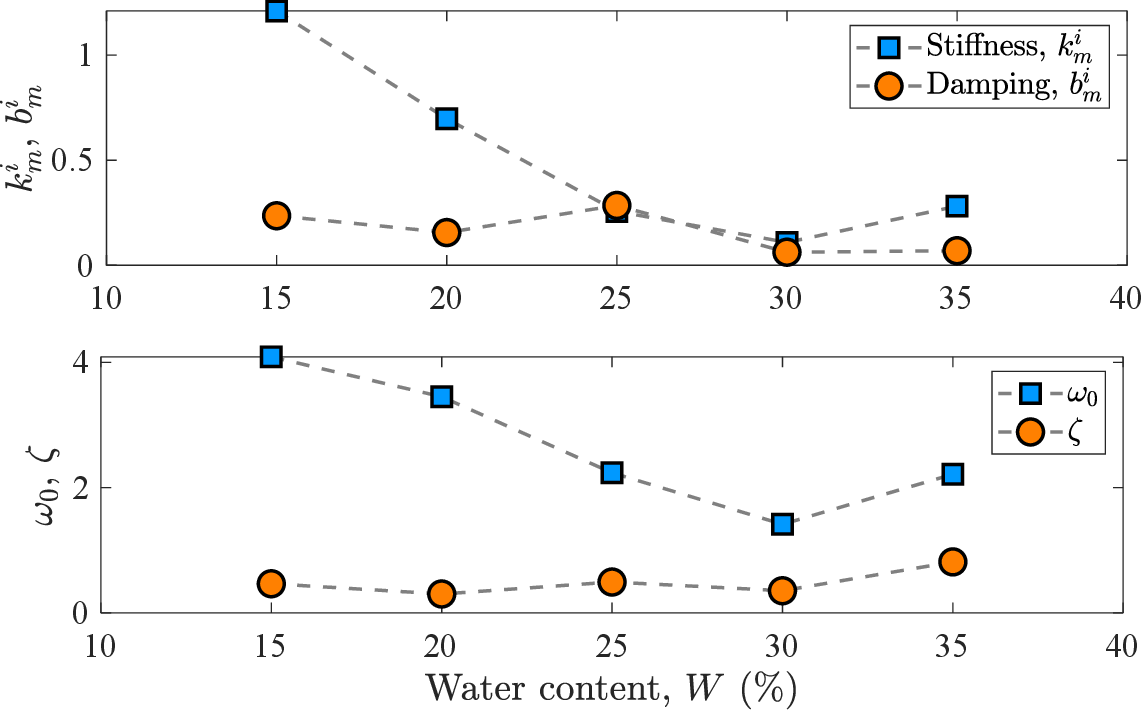}
  \caption{The dependency of the model parameters $k_m^i$, $b_m^i$ and filter parameters $\zeta$, $\omega_0$ on the mud's water content $W$.}
  \label{fig:stiffnessDamping_watercontent}
\end{figure}

We selected experimental results under water content $W=25\%$ to show the model validations. Fig.~\ref{fig:Result_force} shows the model validation results from the perspectives of force profile. The predicted resistive forces by the proposed model matched the experiments accurately over the entire intrusion/withdrawing process. Particularly, by using the visco-elastic spring-damper model, the natural relaxation and decay processes were captured, demonstrating the nonlinearity during the intrusion. Figs.~\ref{fig:Result_velocity} and~\ref{fig:Result_displacement} further show the velocity and displacement profiles respectively for the intruder and the mud. The intruder displacement $z_i$ was obtained by the motion capture systems and its velocity $\dot{z}_i$ was calculated numerically from displacement. The mud displacement $z_m$ and its velocity $\dot{z}_m$ were estimated by the model~\eqref{eqn:spring_damper} with $z_i$ as input. The mud deformation always lagged behind the intruder motion and this created compression of materials, which generated the resistive forces.

We evaluated model estimation performance under different water content conditions by using the identified parameter values in Table~\ref{tab:modelParas}. Fig.~\ref{fig:Error} shows the overall error profiles of multiple experimental runs under various water contents. The plots indicate the mean and one-standard deviation of these error profiles. The estimation error under all tested water content conditions was bounded with $6$~N and the last column in Table~\ref{tab:modelParas} also lists the root mean square error (RMSE) for the average error curves under different water content.

\setcounter{figure}{9}
\begin{figure*}[t!]
	\centering
	\subfigure[]{
		\label{fig:alphaBetaSigma_vel_a}
		\includegraphics[width=2.2 in]{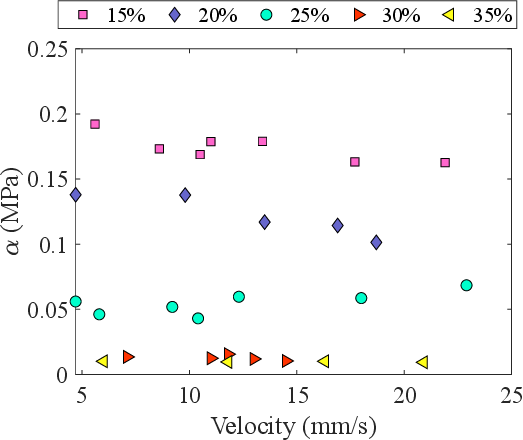}}
	\subfigure[]{
		\label{fig:alphaBetaSigma_vel_b}
		\includegraphics[width=2.15 in]{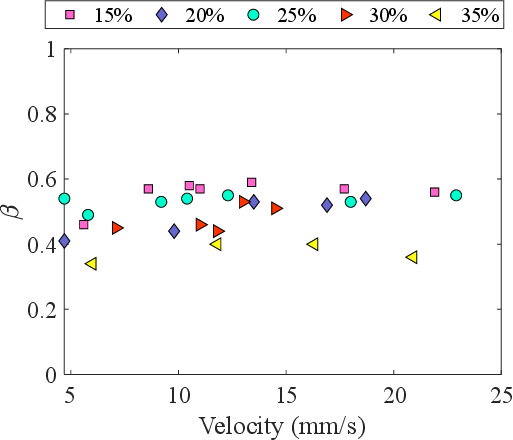}}
	\subfigure[]{
		\label{fig:alphaBetaSigma_vel_c}
		\includegraphics[width=2.3 in]{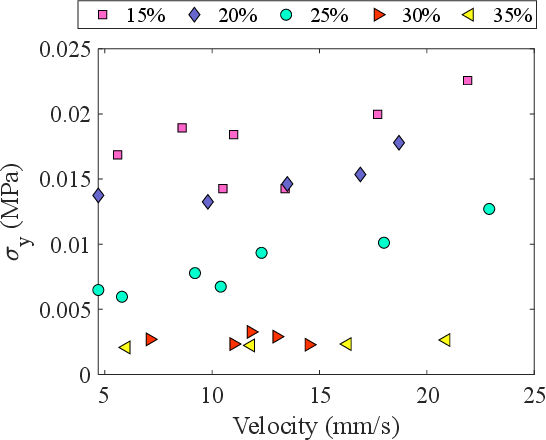}}
	\caption{The estimated parameter values at various intrusion velocity $\dot{z}_i$. (a) Stiffness parameter $\alpha$. (b) Stiffness parameter $\beta$. (c) Mud yielding threshold $\sigma_y$.}
\label{fig:alphaBetaSigma_vel}
\vspace{-2mm}
\end{figure*}

\subsection{Model Parameter Dependency}

The key parameters of interest in the resistive force model include $k_m^i(k_m^w)$, $b_m^i(b_m^w)$, $\alpha$, $\beta$, $\sigma_y$, $\zeta$, and $\omega_0$. By the estimation results of these parameters under various conditions (e.g., water content and intrusion velocity), we obtain useful physical insight into the mud model. It is found that mud stiffness $k_m^i$ ($k_m^w$) and damping coefficient $b_m^i$ ($b_m^w$) are kept consistently under the same experimental condition. Furthermore, the values of parameters $k_m^w$ and $b_m^w$ are kept almost constant even for different water content conditions. We now present the results on how the water content and intrusion velocity influence these model parameters.

Water content variation would change mud stiffness property, that is, both $k_m^i$ and $b_m^i$ values significantly. The top plot in Fig.~\ref{fig:stiffnessDamping_watercontent} shows the dependency of the mud stiffness and damping parameters $k_m^i$ and $b_m^i$ on water content $W$. We observe that above a certain water content level (i.e., $W \leq 30$\%), mud stiffness value decreases significantly with the increasing water content $W$, while damping coefficient values changes only slightly. This observation also agrees with the filter parameters $\omega_0,~\zeta$ as shown in the bottom plot in Fig.~\ref{fig:stiffnessDamping_watercontent}. This is not surprising since $\omega_0 \propto \sqrt{k_m}$ and $\zeta \propto b_m$.

Fig.~\ref{fig:alphabeta_watercontent} further shows the effects of the water content on the stiffness parameters $\alpha$, $\beta$, and yielding threshold $\sigma_y$. As the water content $W$ increases, the stiffness parameter $\alpha$ and yield stress $\sigma_y$ decrease significantly. With a lower water content $W$, mud generates a larger reaction force and the parameter $\beta$ however only decreases sightly as shown in Figs.~\ref{fig:alphabeta_watercontent_a} and~\ref{fig:alphabeta_watercontent_b}. Fig.~\ref{fig:alphabeta_watercontent_c} shows that with a high water content, the value of the yield stress $\sigma_y$ becomes small, implying that the mud becomes flowable but less cohesive and is unable to provide significant suction force before material necking happens.

We used the stable force in the last portion of the sustain regime ($\dot{z}_i=0$) to obtain the estimated $\alpha$ and $\beta$. Therefore, these estimated parameters should not be closely related to the intrusion velocity. Figs.~\ref{fig:alphaBetaSigma_vel_a} and~\ref{fig:alphaBetaSigma_vel_b} show the model parameters $\alpha$ and $\beta$ under various intrusion velocities. For the same water content, the magnitudes of $\alpha$ and $\beta$ are relatively constant. This confirms the above-mentioned model analysis assumption. Yield stress threshold $\sigma_y$ determines when necking happens and the suction force starts reducing. Unlike $\alpha$ and $\beta$, yield stress threshold $\sigma_y$ might be influenced by the withdrawing motion velocity. Fig.~\ref{fig:alphaBetaSigma_vel_c} shows the relationship between $\sigma_y$ and the withdrawal velocity. It is interesting to see that for water contents $W= 15$, $20$, and $25$\%, the value of $\sigma_y$ slightly increases along with the increasing velocity $\dot{z}_i$, while it remains constant (with a small fluctuation) at high water content $W=30$ and $35$\%.

\subsection{Discussions}

\begin{figure}[h!]
	\centering
	\subfigure[]{
		\label{fig:comparisonWetGM_T}
		\includegraphics[width=3.2 in]{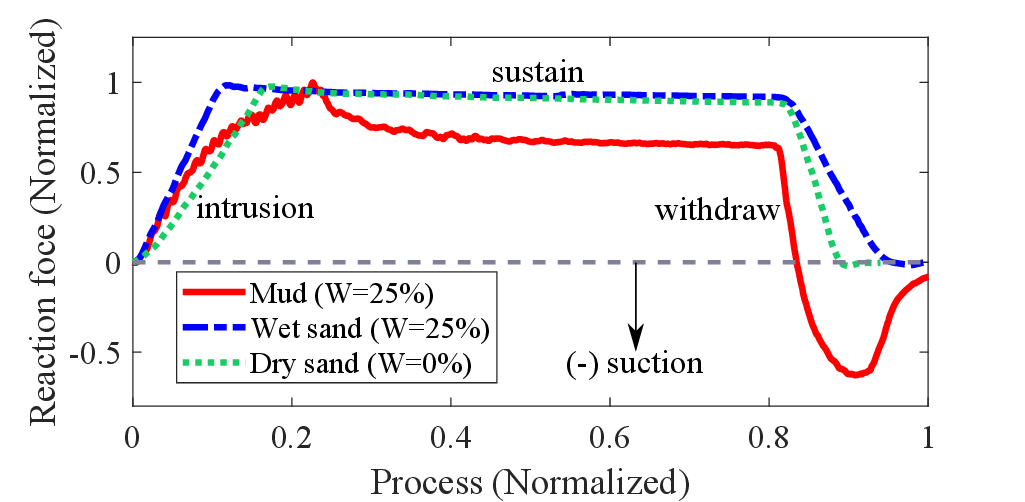}}
	\subfigure[]{
		\label{fig:comparisonWetGM_Z}
		\includegraphics[width=3.2 in]{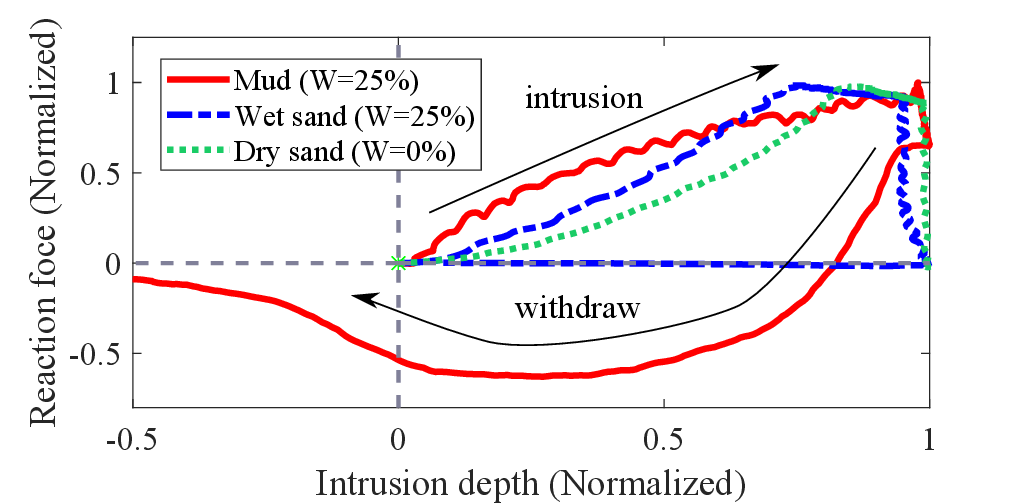}}
	\caption{The reaction force comparisons on mud and dry/wet sand media. The forces and intrusion process were normalized by their maximum magnitudes. (a) The force profiles plotted in time domain. (b) The force profiles plotted over the intrusion displacement $z_i$.}
\label{sandexp}
	\vspace{-3mm}
\end{figure}

To show the significant differences between motion on granular and mud terrains, we conducted intrusion experiments on mud and dry/wet sand. For both mud and wet sand, the water content was set as $W=25$\%. Resistive forces on the intruder were recorded for both terrains. Fig.~\ref{sandexp} shows the comparison experiments of the foot-sand and foot-mud resistive force results. The results were normalized by the maximum values. It is clear that the presence of visco-elaticity (damping) effect was obvious during late intrusion and sustain process. In the sustain regime, the mud resistive force magnitude demonstrated obvious relaxation and decayed to a stable value. However, the dry/wet sand showed a high linearity during the intrusion process and almost no relaxation. Furthermore, no suction force was observed for dry/wet sand. This overall nonlinearity of mud stiffness (reaction force versus intrusion depth) and significant cohesion/suction property made the RFT force models for sand not applicable for mud terrain.

The proposed model considered a parallel, nonlinear spring element for intrusion and sustain process. This is based on the fact of instantaneous force response of the mud due to the action of the external object. Moreover, the nonlinear spring provides necessary steady force when the intruder stops. Otherwise, the force attributed to the visco-elasticity can only correspond to non-zero mud velocity as shown in Fig.~\ref{fig:Result_velocity}. Fig.~\ref{fig:Error} indicates that the model estimated the stable mud resistive force accurately when the intruder was sustained in the mud. However, a large error was observed for all water content conditions in intrusion and withdrawn regimes. This might result from the dynamic effect of moving locomotion in the mud interaction.

From the trend of stiffness coefficient $\alpha$ and yield stress $\sigma_y$ in Figs.~\ref{fig:alphabeta_watercontent_a} and~\ref{fig:alphabeta_watercontent_c}, we can distinguish the water content level. However, the quantitative relationship between water content and these parameters is difficult to predict since it highly depends on mud ingredients. Additionally, the significant drop of the yield stress $\sigma_y$ as the increased water content $W$ implies that inertial effect starts to play a role when the mud materials are sticky with a low water content. For the mud with high water content and flowability, the cohesion of mud rheology becomes insignificant so that yield stress is small and necking easily happens. Therefore, it is possible to use this threshold stress in the model to identify the mud rheological characteristics.

This work found that an increase of water content from $15$ to $35\%$ reduced the resistive force significantly and similar results were reported in~\cite{liu2023adaptation,godon2022insight}. However, the previous findings were empirical while the results in this paper provide modeling analysis with physical interpretation for describing mud rheology. The proposed model uses the foot locomotion velocity as input with tuned parameters to directly estimate resistive force instead of solving constitutive equations to predict shear rate and shear stress as in~\cite{herschel1926measurements,bocquet2009kinetic,bingham1922fluidity,caggioni2020variations,ran2023understanding}, which were difficult to obtain in real time for robotic applications. The proposed modeling approach enables to develop a real-time force estimation by using motion-sensing suites (e.g., inertial measurement unit (IMU)).

Although the proposed model predicted the mud resistive force with a good accuracy, the foot locomotion was considered as a simple vertical intrusion motion. It is necessary to extend this one-dimensional model for applying it to arbitrary three-dimensional robotic locomotion. In addition, we did not consider the effect of foot shape and material properties on foot-mud interactions. We are seeking to overcome these limitations as part of future research.

\section{Conclusions}
\label{Sec:conclusions}

In this paper, we presented a reduced-order model for robotics muddy terrain interactions. We first conducted mud intrusion experiments and highlighted corresponding foot-mud interaction force characteristics. A visco-elasto-plastic analog was used to infer underlying physical mechanism for resistive force estimation. This model took both the intrusion and withdrawing suction into consideration and integrated the robot locomotion into the mud rheological response directly. Through experiments, the proposed model was validated under different water content and locomotion conditions. As an ongoing effort, we plan to extend the 1D resistive force model to 3D locomotion applications for legged robots. Integration of the foot-mud interactions with bipedal dynamics and balance control (e.g.,~\cite{MihalecJBE2022,MihalecTMech2023}) is another ongoing research direction.

\bibliographystyle{IEEEtran}
\bibliography{ChenRef_AIM}

\end{document}